# Testing the Utility of Using Large Language Models to Create Personalized Networks From Therapy Session Transcripts: A Proof of Concept Study

Clarissa W. Ong[1], Hiba Arnaout[2], Kate Sheehan[3], Estella Fox[3], Eugen Owtscharow[2], and Iryna Gurevych[2]

[1]Department of Psychological and Brain Sciences, University of Louisville, U.S.A

[2]Department of Computer Science, TU Darmstadt, Germany

[3]Department of Psychology, University of Toledo, U.S.A

## Author Note

Study materials, including model prompts, annotation examples, model evaluation metrics, and networks generated by our pipeline, are available on OSF at https://osf.io/y3ta6. Tables and figures cited in this report labeled with the letter S can be found in our supplementary materials on the study OSF page. The authors declare no conflict of interest. The authors received no funding from an external source. Corresponding author: Clarissa Ong, University of Louisville, 214 West Barbee St Walk, Life Sciences 109, Louisville, KY 40292, U.S.A. Email: clarissa.ong@louisville.edu

## CRediT Author Statement

CWO: Conceptualization, Data curation, Writing – original draft. HA: Conceptualization, Project administration, Formal analysis, Visualization, Writing – original draft. KS: Data curation, Writing – original draft. EF: Data curation, Writing – original draft. EO: Formal analysis, Visualization, Writing – original draft. IG: Supervision. All authors contributed to Writing – review & editing.

## Abstract

Recent advances in psychotherapy have focused on treatment personalization, such as by selecting treatment modules based on individual networks. However, estimating personalized networks typically requires intensive longitudinal data, which is not always feasible to collect. A solution to increase scalability of network-driven treatment personalization is leveraging large language models (LLMs). In this study, we developed an end-to-end pipeline for automatically generating client networks to support case conceptualization and treatment planning. We annotated 8,028 utterances from 77 therapy transcripts ($N = 6$). In the first stage of the pipeline, we identified clinically relevant processes (binary classification) and their corresponding dimensions (multi-label classification). Then, we introduced a two-step method that grouped the processes into clinically meaningful clusters and generated labels for the clusters. Finally, we generated connections between clusters. Evaluation results generally supported model utility, however, interrater agreement on model performance metrics was inconsistent, ranging from poor to substantial. Qualitative examination of the networks indicated consistency with original study data. Given coherence and interpretability of the generated networks, developing networks from therapy transcripts using LLMs appears to be feasible. Nonetheless, more research is needed to examine whether these networks improve treatment outcomes, including relative to other methods of treatment personalization, such as statistically estimated networks. Potential use cases and limitations of our pipeline are discussed.

**Testing the Utility of Using Large Language Models to Create Personalized Networks From Therapy Session Transcripts: A Proof of Concept Study**

There has been a growing movement away from standardized protocols for diagnoses toward personalized treatment in clinical psychology, with evidence supporting the added value of treatment personalization (Nye et al., 2023). A common method that has been used to tailor treatment to individuals is targeting variables (called *nodes*) in personalized networks, which has produced promising results, even up to 1-year follow-up (Fisher et al., 2019; Levinson et al., 2023). Matching treatment to key network variables is based on network theory, which posits symptoms as interrelated causal mechanisms rather than effects of a latent disorder, such that effectively targeting a symptom in a network has meaningful downstream effects (Bringmann et al., 2022). A metatheoretical framework that builds on network theory (and the behavioral tradition of functional analysis) is a process-based approach to treatment (PBT), which emphasizes person-level clinically relevant processes—defined as patterns of responding related to personally meaningful outcomes, spanning areas such as emotion, cognition, motivation, and sense of self—in their unique context (Hayes et al., 2020). In PBT, processes and the contextual factors influencing them are often depicted in network diagrams—visual representations of clinically relevant variables and the relationships among them—which are then used for case conceptualization and intervention selection (Ong et al., 2025). Networks are continuously revised to monitor clients' response to therapy and adjust treatment.

Despite their utility for treatment personalization, statistically estimated individual-level networks have several limitations with respect to implementation and scalability. First, providing the requisite data can be burdensome for participants, as multiple data points (simulation studies estimate at least 100 observations) are needed for accurate and reliable person-level network estimation (Wright et al., 2025). Thus, estimating networks via statistical methods entails a tradeoff between participant burden and network quality. Second, these networks tend to assess the same set of variables across people, even though symptom heterogeneity—even within the same psychiatric diagnosis—is well-documented (Bryant et al., 2023). As such, statistically estimated personalized networks are not fully personalized. Third, estimating networks minimally require clients to complete surveys, a data collection tool (e.g., survey app), and expertise in analyzing data appropriately or access to such expertise—resources that are not always readily available. Finally, certain statistical assumptions, including stationarity (stability in network characteristics over time), usually need to be met, even though they are rarely achievable in real life (Bringmann et al., 2022), potentially undermining the feasibility of statistical network estimation methods when data are "messy."

As an alternative to these statistical methods, machine learning and large language models (LLMs) could be used to analyze *existing* data, such as therapy sessions (e.g., Eberhardt et al., 2025). Furthermore, LLM-based networks are not limited to survey items, can be automatically generated, and do not entail statistical assumptions about data structure. Already, LLMs have been studied in various psychotherapy and intervention contexts and have demonstrated the capacity to analyze and measure complex constructs (Stade et al., 2024). For example, LLMs have been used to identify therapeutic techniques used by therapists (Hammerfald et al., 2025), track client symptom severity (Abdou et al., 2025), and assess patient engagement in therapy transcripts (Eberhardt et al., 2025). Yet, the application of LLMs to treatment personalization has been limited, with most LLM use cases focusing on symptom detection (e.g., depression screening, diagnostic assessment) and mental health support (e.g., as conversational agents; Guo et al., 2024).

As proposed in the current study, one way to integrate LLMs into treatment personalization is to use them to develop client-specific case formulations in the form of individual networks. Such networks could then be used by clinicians to guide their own case conceptualization and treatment planning, serving a similar function to statistical networks that have been found to lead to improvement in clinical symptoms (Fisher et al., 2019; Levinson et al., 2023). In the study, we developed an end-to-end pipeline for automatically generating personalized networks from therapy session transcripts using LLMs by testing and evaluating different LLM configurations. Our objectives were to (1) examine whether LLMs could be used to generate personalized networks from therapy transcripts and (2) evaluate the potential clinical utility of those networks. Methods used in the current report were not preregistered.

# Method

## Data Source

Our dataset comprised transcripts from 77 1-hour therapy sessions involving six participants diagnosed with major depressive disorder ($n$ = 1) and generalized anxiety disorder ($n$ = 5). The sessions were conducted by two therapists as part of a treatment evaluation study on PBT, which was approved by the local institutional review board (see Ong et al., 2025 for details). The number of sessions per participant ranged from 6 to 16 (mean = 12.8, mode = 15). Therapy occurred between March and October 2023. For online therapy, transcripts were directly downloaded from Webex, a secure teleconferencing platform which offers automated captioning for recorded sessions. In-person sessions were transcribed using OpenAI Whisper (Radford et al., 2023), with speaker diarization performed using the Nvidia NeMo MSDD model. Transcription quality was evaluated using Word Error Rate, where we sampled 2,000 words of utterances and manually assessed their accuracy against the audio file. The accuracy rate was 99% for Whisper and 97% for Webex. After transcription, researchers reviewed the transcripts and removed any identifying information, replacing it with token placeholders (e.g., [HUSBAND], [HOMETOWN], [ORGANIZATION], [YEAR]).

**Procedures**

***Annotating Transcripts to Obtain Ground Truth Data***

The transcript dataset contained 52,799 utterances and a subset of 8,028 utterances were annotated. Annotations were used as "ground truth" data for subsequent model training. Each utterance was annotated by two independent raters. Utterances to be annotated were extracted from the 15-minute segment preceding the final 5 minutes of each therapy session, as this has been hypothesized to be the part of the session where clients are likely to be actively engaged in therapeutic work and would thus contain the most clinically relevant processes for annotation (Auszra et al., 2013). Although variation in sessions meant that these 15 minutes were not always the most active part of a therapy session, the goal was ultimately to accumulate *sufficient* annotated data for subsequent analyses.

Two aspects of each utterance were annotated. First, did the utterance contain a clinically relevant process? This was coded as a binary yes/no variable. Clinically relevant processes were defined as content related to case conceptualization or treatment planning (e.g., presenting concerns, treatment goals, environmental stressors). Content focused on scheduling or small talk was excluded. Second, if clinically relevant, which psychological dimension or level of analysis could it be classified into? The dimensions and levels (hereafter referred to as dimensions collectively for simplicity) used for this annotation were derived from the extended evolutionary meta-model (EEMM), a tool used in PBT to conceptualize psychological functioning in context (Hayes et al., 2020). The EEMM has been previously used to classify treatment mediators (Domhardt et al., 2025) and items in psychological self-report measures (Ciarrochi et al., 2024). The dimension labels used for classification included: *Affect* (how the client feels or responds to feelings about their situation); *Cognition* (how the client thinks and assigns meaning to or responds to thoughts about events); *Attention* (how the client directs or shifts focus during experiences); *Motivation* (reasons for engaging—or not engaging—in behaviors); *Sense of Self* (how the client perceives, conceptualizes, and responds to themselves or their self-concept); *Overt Behavior* (observable actions or inaction taken by the client); *Context/Moderator* (situational factors that are static or difficult to change); *Sociocultural* (interpersonal relationships, including with the therapist, and broader cultural context the client interacts with and is impacted by); and *Biophysiological* (aspects of functioning related to biology and physiology, including interoceptive sensitivity, hormonal changes, sleep, diet, exercise, and chronic health conditions). Dimension was coded as a multi-label variable.

Three therapists—a licensed psychologist and two clinical psychology doctoral students—with training in PBT annotated the transcripts. The raters had previously used the EEMM to classify 1,186 items across a range of psychological self-report measures, achieving interrater reliability (item-total correlations) ranging from 0.66 to 0.88 for all domains except *Context* (*r*s ranged from 0.58 to 0.75; Ciarrochi et al., 2024). Cohen's kappa was used to assess interrater reliability for process identification and each dimension label (Cohen, 1960). Where raters disagreed on whether an utterance contained a process, we erred on over-including processes, such that as long as one rater indicated a process, the utterance was coded as containing a clinically relevant process. Similarly, in cases of disagreement on the dimensions, we included dimensions assigned by both raters.

**Development and Evaluation of Network Construction Pipeline**

To construct personalized networks of clinically relevant processes from therapy transcripts, we designed a three-stage pipeline and used human-in-the-loop evaluation, combining prompt-based LLM reasoning with human evaluation (Schroeder et al., 2025). Each stage employed specialized prompting and analysis procedures tailored to its specific task. To evaluate model output, we adopted a selective and efficiency-oriented evaluation strategy aligned with the structure of our pipeline. Specifically, human-generated annotations were only collected where "ground truth" could be determined more reliably (stage 1), resulting in clearer supervision signals. In contrast, tasks entailing open-ended generation and abstraction (stages 2 and 3), which produced more subjective output and for which we could not generate "ground-truth" data, were evaluated through human ratings. Evaluation metrics used in stages 2 and 3 were developed for the current study based on their respective tasks, considering the proposed function of the networks (i.e., to aid clinical decision making; Fischer et al., 2025). Technical terms related to LLMs are defined in Table 1.

### *Stage 1: Identify and Classify Clinically Relevant Processes*

In stage 1, we identified and categorized clinically relevant processes from therapy session transcripts using a structured prompt-based classification approach. To detect clinically relevant processes from therapy dialogue, we developed a prompt-based approach for context-aware utterance-level classification. Given a single utterance, with the context around it, the prompt instructed the model to first determine whether the utterance reflected a clinically relevant process. If it did, the model was then asked to identify one or more relevant process types from the EEMM dimensions using multi-dimensional labeling. The prompt also included a set of illustrative examples and ended with a classification task for a new utterance. This structured prompt design allowed for consistent and interpretable annotation of clinically relevant processes in transcripts, supporting downstream analysis and modeling.

Our experiments testing the optimal prompt variant included both zero-shot and few-shot settings. In the few-shot setup, we varied the number of in-context examples $K \in \{1,5,10,50,100\}$, sampling balanced examples from the annotated data. For each utterance and value of $K$, we ran the model three times, randomly selecting and shuffling the in-context examples (from a pool of 200 examples randomly sampled from the annotated utterances) to control for selection and order biases (Zhao et al., 2021). The remainder of the annotated utterances not used for in-context prompting were held out for subsequent evaluation. In this stage, we used the LLaMA-3.1-70B-Instruct model (Grattafiori et al., 2024), with a temperature of 0 to ensure deterministic outputs. This model was selected based on its strong performance in recent evaluations of psychological reasoning tasks (Hanafi et al., 2024). In a study comparing 33 LLMs on a range of mental health–related tasks, proprietary models such as GPT-4 and LLaMA 3 demonstrated superior performance. However, given our use of confidential data in this step, we adopted the open-source LLaMA, which could be run locally.

### *Stage 2A: Cluster Processes into Clinical Themes*

After processes were identified and classified, the next step involved organizing them into clinically meaningful clusters and articulating labels (*themes*) for the clusters through a multistep prompt-based clustering approach. Clustering was designed to accommodate the complexity of the transcript data, allowing a single process to be included in multiple clusters when appropriate. For example, “I haven’t started dating again because I don’t know if I’m ready for a relationship” could belong to clusters related to difficulty acting in the presence of uncertainty and avoidance of romantic relationships. This task required the model to recognize underlying functional patterns, rather than merely grouping text by lexical or topical similarity. For example, two seemingly unrelated behaviors, such as “immersing oneself in work” and “engaging in excessive social media scrolling,” may both reflect the broader clinical theme of “distraction from distressing emotions.” We explored two prompt-based clustering strategies using LLaMA-3.1-70B Instruct (temperature = 0). The first was a single-step approach, where the model was prompted to group the processes and assign descriptive cluster labels in a single generation step. The second was a two-step approach (C. Huang & He, 2024), where the model generated a candidate list of cluster labels in the first step, and then assigned each process to one or more of these clusters in the second step. This separation allowed for more controlled and interpretable generation and classification stages.

***Stage 2B: Evaluate Utility of Clinical Themes***

To determine which method (single-step vs. two-step) to build on in stage 3, we applied the single-step and two-step approaches using initial prompt templates and compared their performance on a subset of the data. In two subsequent rounds of refinement, we iteratively improved the prompt by evaluating outputs generated in the previous stage, editing them if necessary, and incorporating revised outputs as gold examples in the next round of refinement. Additionally, we rephrased and added clarifications to the instructions in the prompt based on the outputs and insights from the preceding round. Improvements included refining the prompt template and curating illustrative positive and negative examples from earlier iterations to guide model behavior. In addition, we included the transcript in our model prompt as the model required access to a broader conversational context to make informed judgments.

With respect to evaluating model performance, given the absence of prior research in this specific domain and lack of established criteria for evaluating LLM-generated personalized networks for clinical application, we designed a study-specific evaluation framework considering the goal of developing clinically useful themes to guide case conceptualization and treatment planning. Raters were instructed to compare the model-generated themes to their corresponding input (processes identified in stage 1) and to rate the themes based on several metrics. Themes were generated from 15-min segments of the transcripts to support feasibility of human evaluation, though final reported themes were generated from full transcripts.

We identified the following key evaluation metrics across two broad categories: *Insightfulness* (*Clinical Relevance*, *Novelty*, *Usefulness*) and *Trustworthiness* (*Specificity*, *Coverage*, *Completeness*, *Intrusiveness*, *Redundancy*; see Table 2 for scoring details). *Insightfulness* reflected the extent to which generated themes provided insight into psychological functioning, beyond superficial information provided by the session content, whereas *Trustworthiness* assessed how accurately the network represented the input data and how structurally sound it was for clinical interpretation. Finally, we calculated an overall evaluation score by aggregating all metrics into a weighted linear combination (see Equation 1). The weights were determined based on the relative importance of each metric as ranked by the raters.

$$\begin{aligned} Total\ Score\ =\ & (0.25 \times Clinical\ Relevance) + (0.20 \times Novelty) + (0.15 \times Usefulness) \\ & +\ (0.10 \times Specificity) + (0.10 \times Coverage) + (0.08 \times Completeness) \\ & +\ (0.07 \times Intrusiveness) + (0.05 \times Redundancy)\ (1) \end{aligned}$$

***Stage 3: Generate Explainable Links Between Themes***

In stage 3, we generated directed links accompanied by brief explanations between the themes identified in stage 2 using ensemble prompting strategies (Trad & Chehab, 2025). For each pair of themes (e.g., node A and node B), we asked LLMs to infer the presence or absence of directed connections (i.e., whether A connected to B and/or B connected to A). Each directional decision was made through an ensemble of three models (or model variants) whose outputs were aggregated via majority voting. For every strategy, majority voting determined the final relationship details through the following steps. First, the LLMs identified whether a connection of the same direction (e.g., excitatory or inhibitory) existed. If a majority was reached on the presence of a connection of the same direction, the models proceeded; otherwise, no connection was established. Second, the LLMs indicated the strength of the relationship, selecting the strongest value among the majority outputs. Finally, we randomly sampled an explanation from the subset of outputs that belonged to the majority class and exhibited the strongest connection strength.

In our ensemble framework, we explored three strategies: prompt-based, model-based, and temperature-based ensembles (Trad & Chehab, 2025). In the prompt-based ensemble, we fixed the model (LLaMA-3.1) and temperature (0) and ran three versions of the prompt: zero-shot, one-shot, and few-shot. For the one-shot and few-shot prompts, we incorporated prototypical examples (see Table S1). In the model-based ensemble, we fixed the prompt and varied the model, using LLaMA-3.1-70B-Instruct, Qwen2.5-72B-Instruct (Yang et al., 2025), and GPT-4o-mini (OpenAI, 2024). Finally, in the temperature-based ensemble, we fixed both the prompt and model (LLaMA-3.1) but varied the temperature (0, 0.5, 1.0) to capture outputs ranging from fully deterministic to more diverse. The prompt variants were not compared individually; instead, their outputs were aggregated through majority voting to make a single, more robust decision. We used ensemble learning only in the connection generation step because this last stage of our pipeline involved the most subjective reasoning—inferring directed links between abstract themes—where aggregating multiple LLMs improves reliability. It is also the only stage that operated on high-level, deidentified themes rather than raw transcript data, allowing the inclusion of closed-source models without privacy concerns. Finally, since the number of themes was relatively small compared to the thousands of utterances processed in earlier stages, ensemble inference remained computationally efficient.

For the human evaluation, each rater was presented with three networks generated by the prompt-based, model-based, and temperature-based ensembles respectively across 32 sessions (with overlap to allow calculation of interrater reliability); raters did not know which ensemble produced which network. Clinical themes were fixed across networks and the relationships (or links) between these nodes were generated using distinct ensemble prompting strategies. Raters answered three questions to determine which network best fulfilled predetermined criteria based on our proposed function of the networks (i.e., to inform clinical decision making). Specifically, we evaluated networks on (1) *Clarity* (relatively straightforward to interpret), (2) *Connection Quality* (meaningful and plausible connections among themes), and (3) *Therapeutic Insight* (insight into overall client functioning that might not be obvious from session content). The evaluation questions focused on the quality of the relationships between themes. For each question, the evaluator selected the network that most effectively satisfied the given criteria:

1. *Clarity*: Which network is easier to interpret and navigate in a clinical context?
2. *Connection Quality*: Which network demonstrates more meaningful connections, including the quality of explanations?
3. *Therapeutic Insight*: Which network provides deeper insight into the client and would be more helpful in guiding future therapy sessions?

**Session-Level Network Generation and Evaluation**

We used the LLaMA-3.1-70B-Instruct model to generate all networks ($N = 77$), except for the connection generation step where we used the best-performing model-based ensemble from stage 3. The best-performing method from each stage was used in the final multistep pipeline. Full therapy transcripts were used as input. To assess coherence of model output and consistency between network information and clinical data, we qualitatively examined networks from two randomly selected participants (P11, P13) who completed treatment in the original study. Sessions 1 (intake), 6, 11, and 16 (final session) were selected to represent approximately even intervals between networks.

### Controlling for Known Limitations of LLMs

LLMs are known to exhibit certain limitations (L. Huang et al., 2025; Palikhe et al., 2025; Ranjan et al., 2024), including the potential for biased or inconsistent outputs, hallucination, and lack of transparency in reasoning. To mitigate these risks, several safeguards were implemented in our pipeline. First, we only used open-source models (LLaMA-3.1-70B-Instruct) that could be run locally to process confidential clinical data. Second, except for the temperature-based ensemble in the link generation step, we set the model's temperature to 0 to reduce randomness in generated outputs. Third, prompts included explicit structured instructions and constrained output formats (e.g., JSON schema) to minimize ambiguity and hallucinated information. Finally, identifiers in the transcripts (e.g., gender, ethnicity, occupation) were replaced with neutral placeholders before model input.

## Results

### Process Identification and Dimension Labels

Of the 8,028 utterances, 3,364 were identified as containing clinically relevant processes. Our model correctly identified over 90% of clinically relevant processes (see Figure 1), indicating high accuracy with respect to detecting clinically relevant content. Interrater reliability for process identification (yes/no) was 0.58, while overall agreement on the dimension labels was 0.55, indicating moderate agreement (0.61 is the cutoff for "substantial agreement"; Landis & Koch, 1977). Cohen's kappa values ranged from 0.30 (*Sociocultural*) to 0.85 (*Sense of Self*), indicating significant variation in reliability of dimension coding (see Table S2).

**Network Construction Pipeline**

***Stage 1: Identify and Classify Clinically Relevant Processes***

Precision, recall, and F1 scores for the tasks of (a) identifying whether an utterance contained a process and (b) assigning dimensions to these processes are shown in Figure 1 (see Table S3 for examples). Prompting with examples yielded notable gains (see solid lines in Figure 1) over the zero-shot baseline with no examples (dashed lines), with up to a 15% increase in precision for task (a) and 8% improvement for task (b). For task (b), we found that recall exhibited a slight dip at $K = 10$, before recovering at higher shot counts, which coincided with a corresponding increase in precision. The dip could be due to random rather than intentional selection of in-context examples.

***Stage 2: Cluster Processes into Clinical Themes***

The first iteration of the two-step strategy consistently outperformed the single-step approach across key metrics (maximum score of 3; see Table S4), including *Clinical Relevance* (2.04 vs. 1.94), *Novelty* (1.87 vs. 1.73), and *Usefulness* (1.92 vs. 1.67; see Table 2 for scoring details). Thus, we adopted the two-step strategy in subsequent stages to further refine and evaluate the clustering procedure. The third version of the two-step strategy generally outperformed the initial version, receiving scores of 2.15 for *Clinical Relevance*, 2.25 for *Novelty*, and 2.22 for *Usefulness*. Evaluation results for networks generated by the final (third) version indicated that the networks were generally rated as clinically useful and reliable. However, there was substantial variability in interrater agreement on key evaluation metrics, with Cohen's kappa ranging from 0 (poor) to 0.28 (fair) for *Trustworthiness* metrics and 0.10 (slight) to 0.56 (moderate) for *Insightfulness* metrics, reflecting low agreement between raters on these subjective metrics (see Table S5).

### *Stage 3: Generate Explainable Links Between Themes*

The model-based ensemble method was the most preferred among evaluators (45.9–77.0% of the time), followed by the temperature-based ensemble (18.9–33.8% of the time), while the prompt-based ensemble received the least support (4.1–21.6% of the time). With respect to interrater agreement, Cohen's kappa was 0.61 (substantial) for *Clarity*, 0.40 (moderate) for *Connection Quality*, and 0.10 (slight) for *Therapeutic Insight*.

## Session-Level Networks

Networks from P11 and P13 are presented in Figures 2 and 3 respectively. Further information on participants, including detailed case descriptions, can be found in the original study report (Ong et al., 2025); the same IDs are used to facilitate comparison of findings.

P11's primary presenting concern was insecurity in the context of romantic relationships, especially fear of his partner leaving him and related intense emotional reactions. He also reported worries about school and the future as well as interpersonal stressors related to accepting compliments and social comparisons. His session networks (see Figure 4) highlight themes of difficulty expressing and regulating emotions and a tension between desiring connection while fearing rejection. Over the course of treatment, these themes were further elaborated on. In session 6, fear of rejection shows up again and was linked to a need for control, difficulty trusting others, and hyper-focusing on threats to his relationship. In the second half of treatment, P11 reported continued difficulty with fear of rejection and noted other related struggles, including setting boundaries, establishing a sense of self, and feelings of inadequacy.

The density of connections among the themes suggests multiple positive feedback loops that might be keeping P11 stuck in unhelpful patterns. The final session network illustrates continued struggles with ruminating on negative interpersonal interactions and regulating emotions. However, the network also shows that P11 had begun to accept uncertainty in his relationships, attempt to balance personal boundaries and desire for intimacy, and use values to guide decision making in relationships. Per the network, the latter is hypothesized to reduce difficulties communicating emotional needs.

P13's primary presenting concern was difficulty establishing emotional intimacy with her romantic partner, maintained by all-or-nothing thinking, indecision, and avoidance of distressing situations, including being vulnerable with her partner. P13 also reported significant guilt and self-blame, which negatively impacted her romantic relationship. Her intake network shows a client who is highly guarded in her relationship, contributing to relationship dissatisfaction, rumination, and self-blame (see Figure 5). The rumination and self-blame are supported by indecision and procrastination related to perfectionism. Session 6 appears to have focused specifically on P13's phone use as a strategy for coping with feelings of exhaustion, showing that playing games on her phone provides short-term relief but leads to guilt, potentially related to perfectionistic beliefs about the need to be responsible. Session 11 summarizes skills P13 acquired in treatment, including expressing herself more authentically, letting go of a need for control, and increasing flexibility with respect to rules and boundaries—in contrast to the avoidance of vulnerability and perfectionism seen in the intake network. In the final session, we see that P13 still struggles with being vulnerable in relationships and need for control, reflecting the natural non-linear progression of therapy. At the same time, self-compassion and acceptance of imperfection are present, reinforcing skills highlighted in the session 11 network.

The networks from P11 depict a client whose core struggles center on fear of rejection, uncertainty, inadequacy, and who appeared to gain skills in connecting with values, setting boundaries, and accepting uncertainty over the course of treatment—largely consistent with the case description reported in the original study. Relative to P11, P13 seemed to experience improvement more rapidly, as evidenced by adaptive coping examples as early as session 11, less densely connected networks, and more inhibitory connections at session 16. Nonetheless, themes related to avoidance of vulnerability and perfectionism continued to show up at the end of therapy. This pattern is supported by quantitative data collected via ecological momentary assessment and session surveys, wherein P13 reported the steepest improvement among all participants (Ong et al., 2025). Thus, the session networks largely corroborated qualitative and quantitative data collected in the original study, replicating key themes for each participant.

## Discussion

In the current study, we developed an end–to–end pipeline for automatically generating personalized networks from therapy session transcripts using LLMs. The intended function of these networks is to provide adjunctive support for clinical decision making, potentially as an alternative to statistically estimated networks. Steps in the pipeline include (1) identifying and classifying clinically relevant processes from transcripts, (2) clustering processes into clinically meaningful themes, and (3) generating links between themes. Together, these stages comprise a coherent, transparent, and replicable pipeline for modeling personalized networks of clinically relevant processes from therapy transcripts.

With respect to model performance, our model correctly identified clinically relevant processes over 90% of the time and showed moderate agreement with raters for classifying those processes into EEMM domains. In addition, the model-generated themes were rated favorably, with scores ranging from 2.15 to 2.25 out of 3 on metrics of *Clinical Relevance*, *Novelty*, and *Usefulness*. However, interrater agreement on model performance metrics, including *Novelty* and *Coverage* of clinical themes and *Therapeutic Insight* gleaned from network connections, was inconsistent, indicating variability in how raters evaluated the networks. This result is consistent with interrater agreement for similar subjective tasks reported in a previous study (Mayer et al., 2024). On one hand, low interrater agreement undermines reliability, highlighting a valid concern in terms of whether the networks can be consistently useful across clinicians. On the other hand, the low agreement may reflect naturalistic conditions where clinicians often have divergent perspectives on the same client (Flinn et al., 2015). Moreover, because the intended function of our pipeline is to support clinical decision making, the more important metric may be treatment utility—whether the networks improve treatment outcomes—rather than traditional psychometric indices (Hayes et al., 1987). For example, different interpretations of the same network could still contribute to better outcomes due to equifinality. A qualitative examination of the session-level networks indicated that information provided by the personalized networks was consistent with quantitative and qualitative data reported in the original clinical study. That is, the session-level networks broadly cohered with therapy session content and captured key clinical information, indicating that our pipeline may be able to reliably extract clinically meaningful data from therapy transcripts. Overall, although low agreement between raters on subjective metrics of model performance suggests further evaluation is needed to verify its clinical utility—ideally empirically—our findings support proof of concept for generating coherent and interpretable personalized networks from therapy transcripts. This proof of concept is significant because it shows that generating complex case conceptualizations in the form of network diagrams—which typically requires intensive longitudinal data and advanced

statistics—is feasible using therapy transcripts and LLMs. Thus, our findings, despite being preliminary, signal the potential to build on similar methods to generate networks from real-life clinical data using LLMs.

Based on the network prototypes produced by our pipeline, we see several directions to further improve and evaluate their treatment utility. First, the themes identified by our pipeline, while mapping onto client dialogue, are sometimes overly wordy and contain functional connections within themselves (e.g., “difficulty articulating thoughts and feelings contributes to feelings of inadequacy”), potentially compromising utility of the networks. Hence, a method to more concisely label themes would be beneficial. Second, our pipeline tends to overidentify connections, rendering the networks dense and confusing, even if all the identified connections are plausible. Accordingly, the threshold for generating connections among themes needs to be stricter without introducing too many false negatives. Third, because identifiers were removed from transcripts to maintain participant confidentiality, our networks overlook cultural context, and the ability of our pipeline to integrate cultural considerations into networks is unclear. Ethical considerations around data sharing might continue to make this aspect difficult to test (Arnaout et al., 2026). Finally, given that the use case we envision for this pipeline is to generate session summaries in the form of networks to support clinical decision making, the ultimate test of whether our pipeline meets its goal is in a clinical setting. That is, does using networks generated by our pipeline improve treatment outcomes compared to not using the networks? Empirical investigation of their treatment utility is warranted.

Generating networks using LLMs, such as through our multistep pipeline, may be preferable to statistical estimation of networks given certain advantages that bolster its clinical utility and feasibility. First, our pipeline is easily scalable: the only required input is a therapy session recording and the willingness to have one's therapy session recorded. The latter may be difficult due to privacy concerns but is not a technological or logistical limitation nor is it unique to LLMs. Second, our pipeline more closely approximates a bottom-up approach to case conceptualization, where the network comprises clinically meaningful themes extracted from client utterances, rather than preselected items programmed into a survey. As such, the pipeline begins with the client and is more likely to capture themes unique to them that may not be captured in standardized measures (e.g., "tension between desire for social connection and fear of rejection" for P11, "self-protection through emotional distancing [in relationships]" for P13; see Figures 4 and 5). While it is possible to personalize survey items for clients (e.g., Ong et al., 2025), this approach poses more work on the part of the clinician who has to identify which items are most relevant a priori (most likely in collaboration with their client, taking up therapy time) and customize survey administration, decreasing scalability. Third, the clinically meaningful themes included in the pipeline-generated networks represent *underlying* processes that may not be explicitly stated by clients. As such, our pipeline has the potential to capture patterns outside a client's direct awareness, circumventing the prerequisite of insight for case conceptualization based on self-report. For example, a client may not say that their "fear of abandonment and rejection contributes to anxiety in their relationships," even if their statements are consistent with this pattern (e.g., "I don't want to get too close in case I get hurt"). In this way, the themes add value to rather than merely rehash content from therapy transcripts.

**Potential Applications of the Pipeline**

Given these advantages, various applications of our pipeline may support clinical practice, training, and research. With respect to clinical work, the networks could be used to provide postsession feedback to clinicians on what themes were most salient to their client in the session. Especially considering that clinicians appear to prefer to use AI tools to support clinician-driven decision making rather than make decisions for them (Fischer et al., 2025), simply providing a functional session summary gives clinicians latitude in how to proceed with treatment. For example, a clinician might miss an underlying theme, and the clinician could then update their hypothesized case conceptualization with the network-identified theme. Over the course of treatment, the networks could be used to supplement clinicians' case conceptualizations, serving as a "back-up" observer of each session, decreasing the likelihood of missing key or recurring themes. Ultimately, we see these personalized networks as a tool to *enhance* case conceptualization and treatment planning that is still initiated and executed by the clinician, not as a replacement for clinical decision making. Indeed, given the potential of LLMs to hallucinate (L. Huang et al., 2025), it is important that these networks are not assumed to be truth and verified against clinician judgment and client self-report as appropriate.

With respect to training, trainees could use the networks as feedback to refine and test their developing case conceptualization skills. For instance, trainees could compare their networks to the model-generated networks and critically evaluate discrepancies, clarifying why they might have missed a theme. In a training context, the purpose could be honing critical thinking skills more so than searching for an objectively accurate model (e.g., providing clinical rationale for why the model is "wrong"). As such, accuracy may be less of a priority for training purposes. After all, fully appreciating a client's history, identities, lived experience, and current circumstances is a Sisyphean task—the moment clients leave the therapy room, a whole fresh set of variables is introduced into their lives. A clinician's goal is not necessarily to be correct; it is to help their clients. Hence, using the networks as feedback to develop therapeutic skills and clinical decision making that will strengthen trainees' ability to support their clients' growth and well-being seems more meaningful than as a way to check for "correct" answers.

For research, the biggest contribution of our pipeline is scalability. Statistical power is a perennial consideration in quantitative studies, with the desired sample size (e.g., number of observations per person for idiographic models) generally increasing as analyses become more complex (Wright et al., 2025). Yet, collecting usable data is resource-intensive, in terms of time, money, and human effort. Our pipeline could provide an alternative to statistical network estimation where resources are limited, if the networks demonstrate clinical utility. In addition, tasks in the pipeline could increase scalability of qualitative analysis of clinical data, enabling existing textual data to be readily converted into meaningful variables (e.g., frequency of EEMM dimensions covered in therapy sessions), increasing the scope of personalized treatment research.

**Limitations**

While our study provides promising insights and suggests potential practical applications, several limitations should be acknowledged. First, we used a single LLM (LLaMA-3.1-70B-Instruct) for the majority of the network generation pipeline, and different results may be achieved if more models were included for performance comparison. Second, although our approach incorporated multiple safeguards against known limitations of LLMs, such as bias and hallucination, fully eliminating these issues is beyond the scope of this work. Therefore, networks should be used as hypothesis-generating tools rather than "true" answers. Third, the dataset used to develop and test our pipeline was relatively small and drawn from one study. Although sufficient for our initial proof of concept evaluation, scaling to a larger and more diverse dataset could strengthen model performance and generalizability. Fourth, our model input did not include nonverbal behaviors, which play an important role in the therapeutic process (Del Giacco et al., 2020). As such, networks are constrained by verbally mediated information and necessarily discount meaningful nonverbal cues that would otherwise be accessible to the clinician in the therapy room. This is another reason why we advocate using networks as a clinical support tool, not to replace clinical observation and judgment. Fifth, there was low interrater agreement on several evaluation metrics, suggesting that the networks were not consistently perceived as useful by human raters. Although this is undoubtedly a problem with respect to reliability, we developed the pipeline to provide *useful* support to clinicians. As such, our primary metric for the merit of the networks is whether they improve treatment outcomes (Hayes et al., 1987), and the pipeline could be reliably useful even if not reliably accurate. Empirical investigations are needed to determine its treatment utility.

## Conclusion

In this work, we introduced an end-to-end pipeline that creates session-level personalized networks to support clinical decision making. Using therapy transcripts, we employed prompt-based methods in multistep network generation: first identifying and classifying clinically relevant processes, then clustering them into clinical meaningful themes, and finally generating plausible connections between these themes. Despite variable agreement from raters on performance metrics, the networks generally cohered with session notes, clinical assessment, and self-report data from the original study, suggesting convergent validity with its source material. This work provides a proof of concept for using LLMs to automatically construct psychologically meaningful representations of therapy sessions from their transcripts. Nonetheless, more research is needed to evaluate the clinical utility of these networks. Ultimately, the intended application of our pipeline and its resulting networks is to provide a scalable tool to support treatment personalization and improve treatment outcomes.

## References

Abdou, M., Sahi, R. S., Hull, T. D., Nook, E. C., & Daw, N. D. (2025). Leveraging large language models to estimate clinically relevant psychological constructs in psychotherapy transcripts. *medRxiv*, 2025–03.

Arnaout, H., Goel, A., Schwartz, H. A., Eberhardt, S. T., Atzil-Slonim, D., Doherty, G., Schwartz, B., Lutz, W., Althoff, T., Choudhury, M. D., Jamalabadi, H., Shah, R. S., Plaza-del-Arco, F. M., Hovy, D., Liakata, M., & Gurevych, I. (2026). *Responsible evaluation of AI for mental health*. 2602.00065.

Auszra, L., Greenberg, L. S., & Herrmann, I. (2013). Client emotional productivity—Optimal client in-session emotional processing in experiential therapy. *Psychotherapy Research*, *23*(6), 732–746. https://doi.org/10.1080/10503307.2013.816882

Bringmann, L. F., Albers, C., Bockting, C., Borsboom, D., Ceulemans, E., Cramer, A., Epskamp, S., Eronen, M. I., Hamaker, E., Kuppens, P., Lutz, W., McNally, R. J., Molenaar, P., Tio, P., Voelkle, M. C., & Wichers, M. (2022). Psychopathological networks: Theory, methods and practice. *Behaviour Research and Therapy*, *149*, 104011. https://doi.org/10.1016/j.brat.2021.104011

Bryant, R. A., Galatzer-Levy, I., & Hadzi-Pavlovic, D. (2023). The heterogeneity of posttraumatic stress disorder in DSM-5. *JAMA Psychiatry*, *80*(2), 189–191. https://doi.org/10.1001/jamapsychiatry.2022.4092

Ciarrochi, J., Hernández, C., Hill, D., Ong, C., Gloster, A. T., Levin, M. E., Yap, K., Fraser, M. I., Sahdra, B. K., Hofmann, S. G., & Hayes, S. C. (2024). Process-based therapy: A common ground for understanding and utilizing therapeutic practices. *Journal of Psychotherapy Integration*, *34*(3), 265–290. https://doi.org/10.1037/int0000348

Cohen, J. (1960). A coefficient of agreement for nominal scales. *Educational and Psychological Measurement*, *20*(1), 37–46. https://doi.org/10.1177/001316446002000104

Del Giacco, L., Anguera, M. T., & Salcuni, S. (2020). The action of verbal and non-verbal communication in the therapeutic alliance construction: A mixed methods approach to assess the initial interactions with depressed patients. *Frontiers in Psychology*, *Volume 11-2020*. https://www.frontiersin.org/journals/psychology/articles/10.3389/fpsyg.2020.00234

Domhardt, M., Mennel, V., Angerer, F., Grund, S., Mayer, A., Büscher, R., Sander, L. B., Cuijpers, P., Terhorst, Y., & Baumeister, H. (2025). Processes of change in digital interventions for depression: A meta-analytic review of cognitive and behavioral mediators. *Behaviour Research and Therapy*, *189*, 104735. https://doi.org/10.1016/j.brat.2025.104735

Eberhardt, S. T., Vehlen, A., Schaffrath, J., Schwartz, B., Baur, T., Schiller, D., Hallmen, T., André, E., & Lutz, W. (2025). Development and validation of large language model rating scales for automatically transcribed psychological therapy sessions. *Scientific Reports*, *15*(1), 29541. https://doi.org/10.1038/s41598-025-14923-y

Fischer, L., Mann, P. A., Nguyen, M.-H. H., Becker, S., Khodadadi, S., Schulz, A., Edwin Thanarajah, S., Repple, J., Hahn, T., Reif, A., Salamikhanshan, A., Kittel-Schneider, S., Rief, W., Mulert, C., Hofmann, S. G., Dannlowski, U., Kircher, T., Bernhard, F. P., & Jamalabadi, H. (2025). AI for mental health: Clinician expectations and priorities in computational psychiatry. *BMC Psychiatry*, *25*(1), 584. https://doi.org/10.1186/s12888-025-06957-3

Fisher, A. J., Bosley, H. G., Fernandez, K. C., Reeves, J. W., Soyster, P. D., Diamond, A. E., & Barkin, J. (2019). Open trial of a personalized modular treatment for mood and anxiety. *Behaviour Research and Therapy*, *116*, 69–79. https://doi.org/10.1016/j.brat.2019.01.010

Flinn, L., Braham, L., & das Nair, R. (2015). How reliable are case formulations? A systematic literature review. *British Journal of Clinical Psychology*, *54*(3), 266–290. https://doi.org/10.1111/bjc.12073

Grattafiori, A., Dubey, A., Jauhri, A., Pandey, A., Kadian, A., Al-Dahle, A., Letman, A., Mathur, A., Schelten, A., Vaughan, A., Yang, A., Fan, A., Goyal, A., Hartshorn, A., Yang, A., Mitra, A., Sravankumar, A., Korenev, A., Hinsvark, A., … Ma, Z. (2024). *The Llama 3 herd of models*. arXiv preprint arXiv:2407.21783. https://arxiv.org/abs/2407.21783

Guo, Z., Lai, A., Thygesen, J. H., Farrington, J., Keen, T., & Li, K. (2024). Large language models for mental health applications: Systematic review. *JMIR Ment Health*, *11*, e57400. https://doi.org/10.2196/57400

Hammerfald, K., Schmidt, F., Vlassov, V., Haaland Jahren, H., & Solbakken, O. A. (2025). Leveraging large language models to identify microcounseling skills in psychotherapy transcripts. *Psychotherapy Research*, 1–19. https://doi.org/10.1080/10503307.2025.2539405

Hanafi, A., Saad, M., Zahran, N., Hanafy, R. J., & Fouda, M. E. (2024). A comprehensive evaluation of large language models on mental illnesses. *arXiv Preprint arXiv:2409.15687*.

Hayes, S. C., Hofmann, S. G., & Ciarrochi, J. (2020). A process-based approach to psychological diagnosis and treatment:The conceptual and treatment utility of an extended evolutionary meta model. *Clinical Psychology Review*, *82*, 101908. https://doi.org/10.1016/j.cpr.2020.101908

Hayes, S. C., Nelson, R. O., & Jarrett, R. B. (1987). The treatment utility of assessment: A functional approach to evaluating assessment quality. *American Psychologist*, *42*(11), 963–974. https://doi.org/10.1037/0003-066X.42.11.963

Huang, C., & He, G. (2024). Text clustering as classification with LLMs. *arXiv Preprint arXiv:2410.00927*.

Huang, L., Yu, W., Ma, W., Zhong, W., Feng, Z., Wang, H., Chen, Q., Peng, W., Feng, X., Qin, B., & Liu, T. (2025). A survey on hallucination in large language models: Principles, taxonomy, challenges, and open questions. *ACM Trans. Inf. Syst.*, *43*(2). https://doi.org/10.1145/3703155

Landis, J. R., & Koch, G. G. (1977). The measurement of observer agreement for categorical data. *Biometrics*, *33*(1), 159–174. https://doi.org/10.2307/2529310

Levinson, C. A., Williams, B. M., Christian, C., Hunt, R. A., Keshishian, A. C., Brosof, L. C., Vanzhula, I. A., Davis, G. G., Brown, M. L., Bridges-Curry, Z., Sandoval-Araujo, L. E., & Ralph-Nearman, C. (2023). Personalizing eating disorder treatment using idiographic models: An open series trial. *Journal of Consulting and Clinical Psychology*, *91*(1), 14–28. https://doi.org/10.1037/ccp0000785

Mayer, T., Warikoo, N., Eliassaf, A., Atzil-Slonim, D., & Gurevych, I. (2024). Predicting client emotions and therapist interventions in psychotherapy dialogues. *Proceedings of the 18th Conference of the European Chapter of the Association for Computational Linguistics (Volume 1: Long Papers)*, 1463–1477.

Nye, A., Delgadillo, J., & Barkham, M. (2023). Efficacy of personalized psychological interventions: A systematic review and meta-analysis. *Journal of Consulting and Clinical Psychology*, *91*(7), 389–397. https://doi.org/10.1037/ccp0000820

Ong, C. W., Sheehan, K., Mann, A. J. D., & Fox, E. (2025). Examining the effects of process-based therapy: A multiple baseline study. *Journal of Contextual Behavioral Science*, *35*, 100875. https://doi.org/10.1016/j.jcbs.2025.100875

OpenAI, G. (2024). 4o mini: Advancing cost-efficient intelligence, 2024. *URL: Https://Openai. Com/Index/Gpt-4o-Miniadvancing-Cost-Efficient-Intelligence*.

Palikhe, A., Yu, Z., Wang, Z., & Zhang, W. (2025). *Towards transparent AI: A survey on explainable large language models*. https://arxiv.org/abs/2506.21812

Radford, A., Kim, J. W., Xu, T., Brockman, G., Mcleavey, C., & Sutskever, I. (2023). Robust speech recognition via large-scale weak supervision. In A. Krause, E. Brunskill, K. Cho, B. Engelhardt, S. Sabato, & J. Scarlett (Eds.), *Proceedings of the 40th International Conference on Machine Learning* (Vol. 202, pp. 28492–28518). PMLR. https://proceedings.mlr.press/v202/radford23a.html

Ranjan, R., Gupta, S., & Singh, S. N. (2024). A comprehensive survey of bias in LLMs: Current landscape and future directions. *arXiv Preprint arXiv:2409.16430*.

Schroeder, H., Roy, D., & Kabbara, J. (2025). Just put a human in the loop? Investigating LLM-assisted annotation for subjective tasks. In W. Che, J. Nabende, E. Shutova, & M. T. Pilehvar (Eds.), *Findings of the Association for Computational Linguistics: ACL 2025* (pp. 25771–25795). Association for Computational Linguistics. https://doi.org/10.18653/v1/2025.findings-acl.1323

Stade, E. C., Stirman, S. W., Ungar, L. H., Boland, C. L., Schwartz, H. A., Yaden, D. B., Sedoc, J., DeRubeis, R. J., Willer, R., & Eichstaedt, J. C. (2024). Large language models could change the future of behavioral healthcare: A proposal for responsible development and evaluation. *Npj Mental Health Research*, *3*(1), 12. https://doi.org/10.1038/s44184-024-00056-z

Trad, F., & Chehab, A. (2025). To ensemble or not: Assessing majority voting strategies for phishing detection with large language models. In A. Bennour, A. Bouridane, S. Almaadeed, B. Bouaziz, & E. Edirisinghe (Eds.), *Intelligent Systems and Pattern Recognition* (pp. 158–173). Springer Nature Switzerland.

Wright, A. G. C., Scharf, F., & Zimmermann, J. (2025). Minimum sampling recommendations for applied ambulatory assessment. *Psychological Assessment*, *37*(10), 466–478. https://doi.org/10.1037/pas0001408

Yang, A., Li, A., Yang, B., Zhang, B., Hui, B., Zheng, B., Yu, B., Gao, C., Huang, C., Lv, C., Zheng, C., Liu, D., Zhou, F., Huang, F., Hu, F., Ge, H., Wei, H., Lin, H., Tang, J., … Qiu, Z. (2025). Qwen3 Technical Report. *CoRR*, *abs/2505.09388*. https://doi.org/10.48550/ARXIV.2505.09388

Zhao, Z., Wallace, E., Feng, S., Klein, D., & Singh, S. (2021). Calibrate before use: Improving few-shot performance of language models. *International Conference on Machine Learning*, 12697–12706.

**Table 1**

***Definitions of Key Technical Terms Related to Large Language Models***

| Concept | Definition / Description |
|---|---|
| Large language models (LLMs) | Computer programs designed to understand and generate human language. They are trained on very large collections of text (e.g., books, websites, articles) and learn patterns in how words and sentences are used. LLMs can perform many language-related tasks, such as answering questions, summarizing text, translating between languages, or having a conversation. They are based on a type of artificial intelligence called machine learning, which allows them to improve by learning from examples, rather than being explicitly programmed. |
| Open-source LLMs | Language models whose code and trained weights are publicly available for anyone to use, modify, and improve. They can be run locally, allowing organizations to process sensitive data, such as mental health information, without sending it to external servers. This makes them ideal for applications that require privacy, security, and full control over data. |
| In-context learning | The ability of an LLM to learn a task by being given instructions and examples regarding that task directly in the prompt, without updating its parameters. Instead, the model uses the provided examples and structure in the prompt to adapt its response. |
| Few-shot prompting | A technique used with LLMs in which a small number of input–output examples are provided within the prompt to guide the model in performing a new task. Typically, this involves presenting examples of the desired behavior or pattern before asking the model to continue or respond in the same way. Few-shot prompting is a specific case of in-context learning. |
| F1 score | A measurement used to evaluate how well a model performs on classification tasks, that is, tasks where the model must assign items to categories (e.g., “True” or “False”). It combines two key aspects of performance: (1) *Precision*: how many items the model labeled as positive were correct, (2) *Recall*: how many positive items the model was able to identify. The F1 Score is the harmonic mean of precision and recall, which means it balances the two. It ranges from 0 (worst) to 1 (best). Mathematically, it is defined as: $$F1\ Score = 2 \cdot \frac{Precision \cdot Recall}{Precision + Recall}$$ This metric is widely used in machine learning and natural language processing to assess classification accuracy in a more nuanced way than overall correctness alone. |

**Table 2**

***Metrics Used to Evaluate Clinically Meaningful Clusters in Stage 2B***

| Category | Metric | Description | Scoring |
|---|---|---|---|
| Insightfulness | Clinical Relevance | Are the identified clinical themes meaningful to the client's functioning, well-being, and goals? | 1: Not Meaningful<br>2: Moderately Meaningful<br>3: Highly Meaningful |
| | Novelty | Does the network reveal any new or unexpected insights? | 1: Low or No Novelty<br>2: Some Novelty<br>3: High Novelty |
| | Usefulness | Does the network support clinical decision making? | 1: Low Usefulness<br>2: Moderate Usefulness<br>3: High Usefulness |
| Trustworthiness | Specificity | Is the network too broad or sufficiently specific to be clinically useful? | 1: Too Broad<br>2: Moderately Specific<br>3: Highly Specific |
| | Coverage | Is the network missing clinical themes? | 1: Incomplete<br>2: Partial Coverage<br>3: Complete |
| | Completeness | Is the network missing nodes (processes) from the input? | Computed automatically† |
| | Intrusiveness | Is there an intruder (irrelevant) node in the cluster? | 1: Many Intruders<br>2: Few Intruders<br>3: No Intruders |
| | Redundancy | Are there any clinically relevant themes that are redundant in the network? | 1: High Redundancy<br>2: Moderate Redundancy<br>3: Low or No Redundancy |

†Completeness measures whether all processes from the input have been incorporated into the network and was computed using simple string matching.

**Figure 1**

***Model Performance on (a) Process Identification and (b) Dimension Classification***

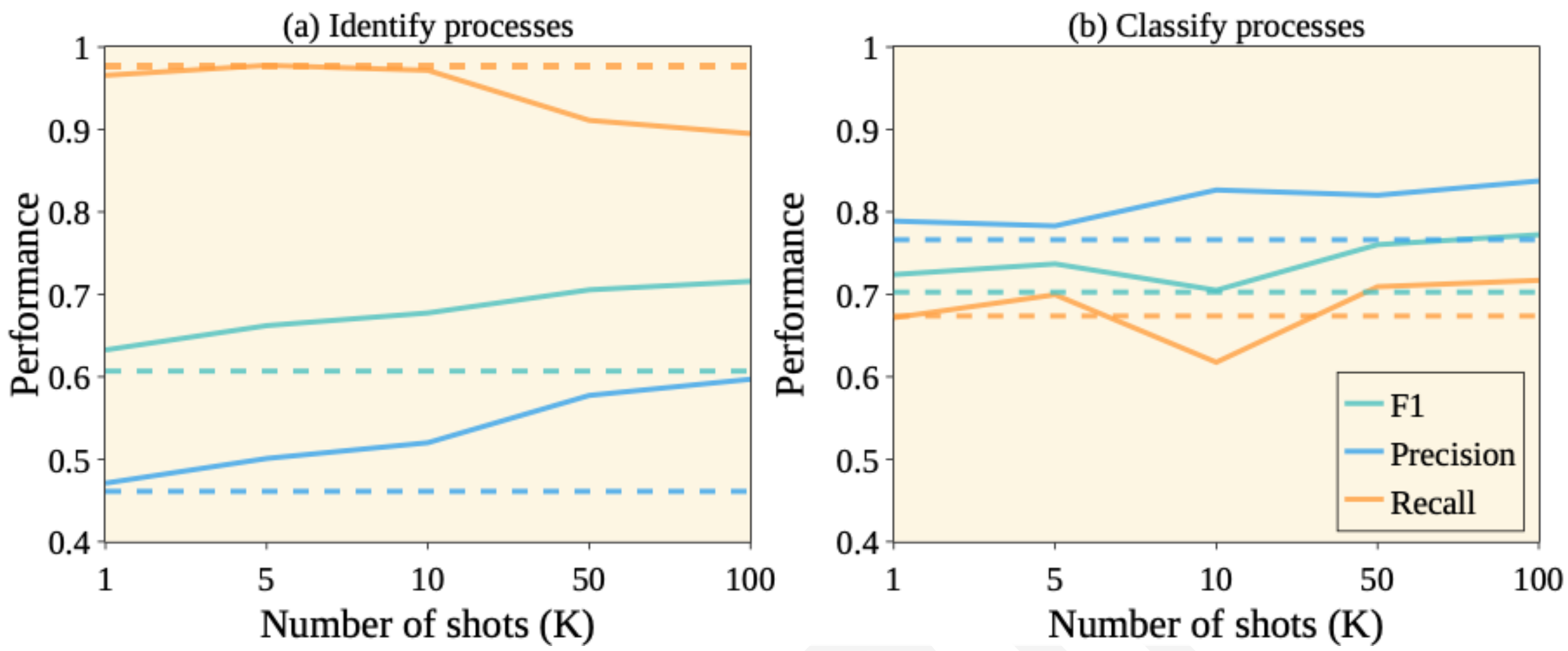

*Note.* The solid lines represent evaluation results as number of shots increases (up to $K = 100$); the dashed lines show results of the zero-shot prompt ($K = 0$). Precision refers to positive predictive power (accuracy of positive predictions), and recall refers to sensitivity (ability to identify positive instances). The F1 score takes into account both precision and recall. See Table 1 for detailed definitions of large language model terms.

**Figure 2**

***P11 Networks Generated From Multistep Pipeline***

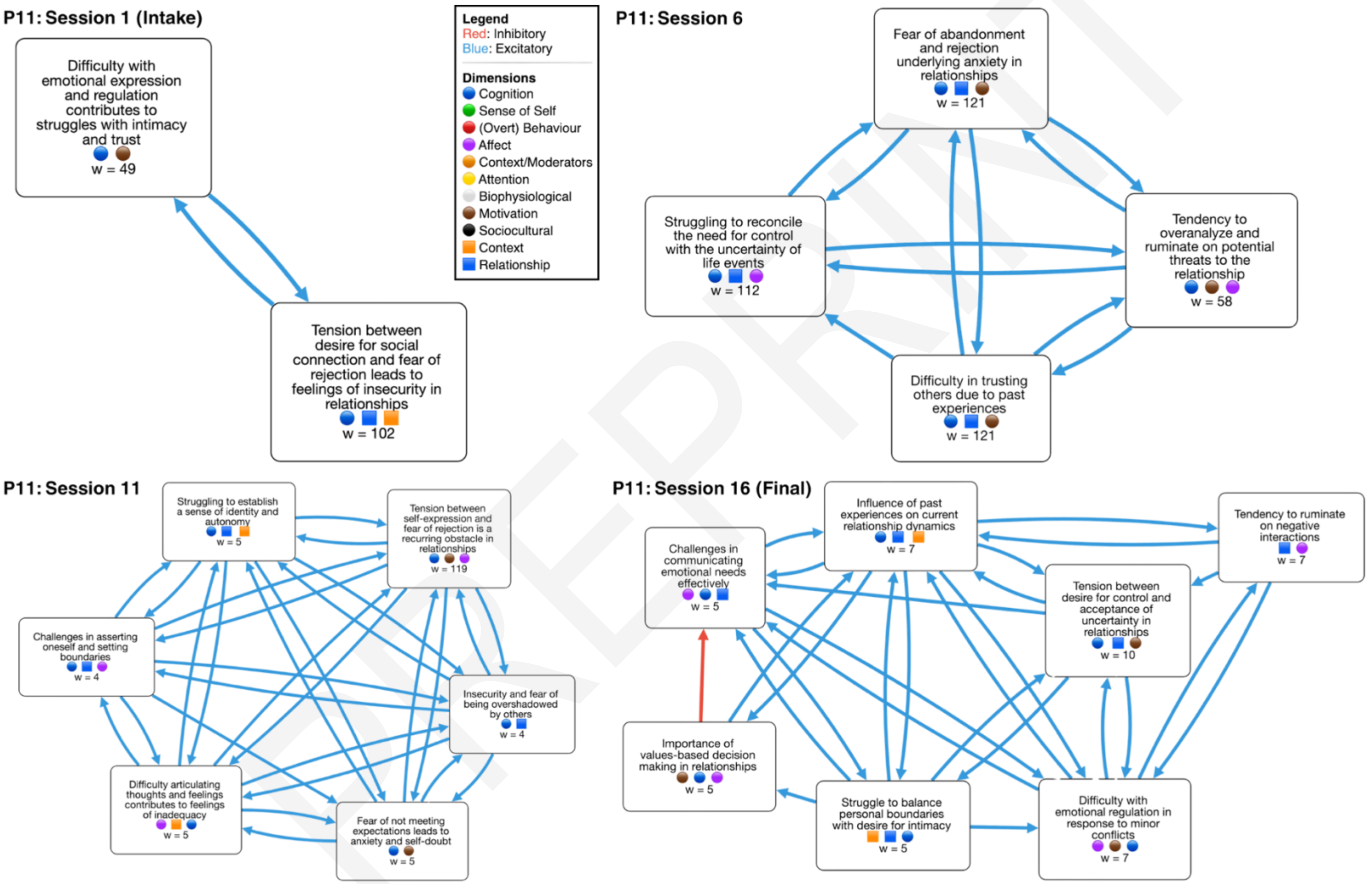

*Note.* w = weight of the cluster (i.e., number of processes included in the cluster).

**Figure 3**

***P13 Networks Generated From Multistep Pipeline***

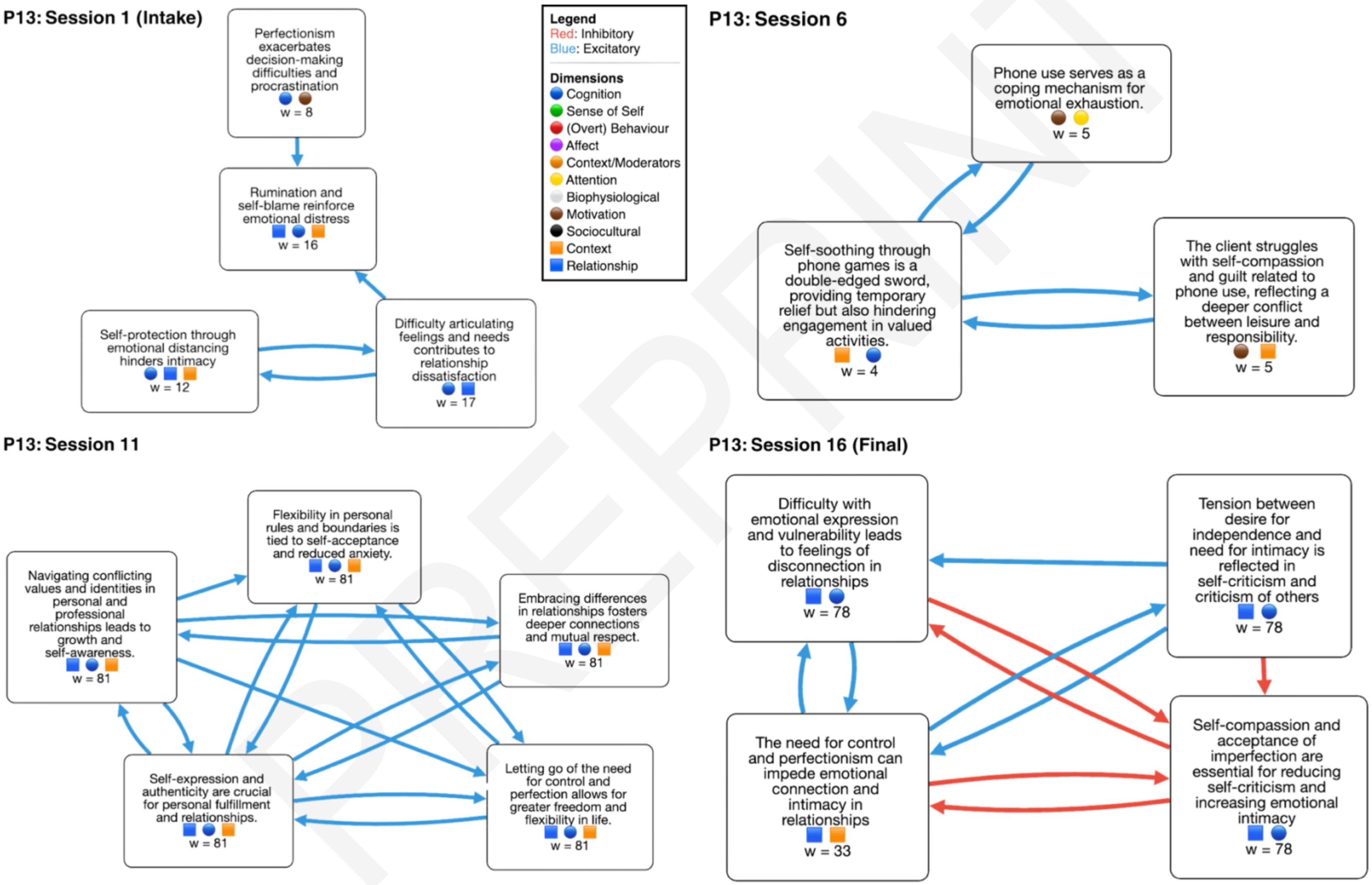

*Note.* w = weight of the cluster (i.e., number of processes included in the cluster).